\documentclass[letterpaper, 10 pt, conference]{ieeeconf}  

\IEEEoverridecommandlockouts                              

\usepackage{graphics}  
\usepackage{epsfig}    
\usepackage{amsmath}   
\usepackage{bm}
\usepackage{mathtools}
\usepackage{commath}
\usepackage{color}
\usepackage{soul}
\usepackage{amssymb}
\usepackage{comment}
\usepackage{xfrac}
\usepackage{nicefrac}

\usepackage{hyperref}
\usepackage{url}
\usepackage[noadjust]{cite}
\usepackage{xcolor}
\usepackage[percent, abs]{overpic}

\usepackage[font=small]{caption}
\usepackage[font=small]{subcaption}

\usepackage{algorithm}
\usepackage{algorithmic}

\usepackage[bottom]{footmisc}

\usepackage{graphicx}
\graphicspath{ {C:/Users/work/Desktop/PhD Work/latex work/figures/} }
\renewcommand{\baselinestretch}{0.96}

\title{\LARGE \bf
Laser Powered Harvesting System for Table-Top Grown Strawberries
}

\author{Mohamed Sorour, Pål Johan From 
\thanks{Robotics Group, Faculty of Science and Technology, Norwegian University of Life Sciences (NMBU), 1432 Ås, Norway.\newline%
{\tt\small msorour@nmbu.no}\newline \indent%
}}

\begin{document}

\maketitle
\thispagestyle{empty}
\pagestyle{empty}

\begin{abstract}
In this paper, a novel tool prototype for harvesting table-top grown strawberries is presented. With robustness against strawberry localization error of $\pm15mm$ and average cycle time of $8.02$ seconds at $50$\% of maximum operational velocity, it provides a promising contribution towards full automation of strawberry harvesting. In addition, the tool has an overall fruit-interacting width of $35mm$ that greatly enhances reach-ability due to the minimal footprint. A complete harvesting system is also proposed that can be readily mounted to a mobile platform for field tests. An experimental demonstration is performed to showcase the new methodology and derive relevant metrics. 


\end{abstract}


\section{Introduction}
On account of the seasonal nature of manual harvesting \cite{Nolte2017,uk_gouv}, increasing demand for labours \cite{workforce_in_agriculture_world_bank,nfu_labour_availability_issues} in competitive industrial sectors, as well as ageing farming society \cite{Duckett2018}, the need for harvesting automation is well established and justified. Such unmet need, results in elevated cost of living \cite{Cassey2018} due to the higher costs incurred to secure labour, and lesser supply of crop, damaged for missing the optimal harvesting window. Harvesting robots developed so far, on the other hand, have an average success rate as low as $66\%$ \cite{Bac2014}, subject to drastic drop in cluttered environment, mostly due to bulky harvesting tools \cite{Kootstra2021}.

\begin{figure}[t!]
    \centering
    \includegraphics[scale = 1.1]{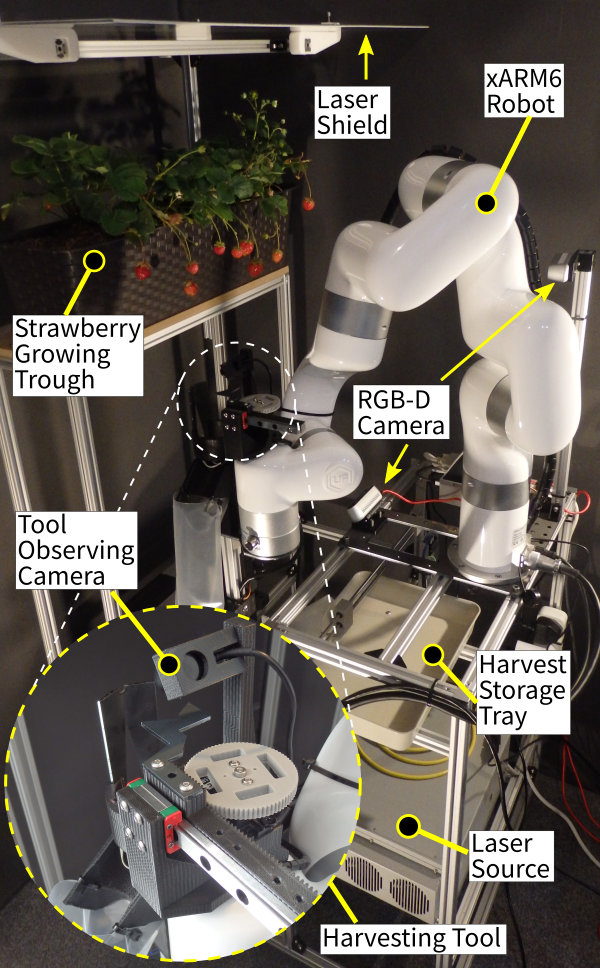}
    \captionsetup{aboveskip=10pt}
    \captionsetup{belowskip=-10pt}
    \caption{Harvesting setup simulating an actual table-top strawberry growing system in poly-tunnels.}
    \label{figure_setup_overview}
\end{figure}

Automated fruit harvesting essentially requires a tool that can: (1) capture the fruit, and (2) detach it. To capture, grasping is conventionally employed in the literature, harvesting apples \cite{DE-AN2011,Silwal2017,Onishi2019}, plums \cite{Jasper2021}, kiwi \cite{MU2020}, tomatoes \cite{Feng2015}, sweet pepper \cite{Arad2020,Bac2017}, and strawberries \cite{Xiong2019}. Mechanical grasping mostly results in a bulky tool interacting with the produce, hindering as such reach-ability. Employing vacuum suction instead \cite{Baeten2008,Tanigaki2008,Hayashi2009,Hayashi2014,HU_2022}, reduces the tool size, but lead to complexity in finding the correct spot on the produce to apply, as well as being force-full. To detach, sharp mechanical cutter is by far the most commonly used method, enlarging the harvesting tool size, since the driving source (motor) and transmission must be very close to the cutting tool. Non-conventional cutting means are rare in the literature, featuring an oscillating blade to cut sweet pepper in \cite{Lehnert2017}, and a thermal cutting device in \cite{vanHenten2002,Bachche2013} for cucumber and pepper respectively. Laser beam is used in \cite{Liu2008,Liu2011} to showcase the potential for cutting tomato peduncles, and in \cite{Heisel2002, Mathiassen2006, Coleman2021, Nadimi2021} for weed control. Despite the non-conventional cutting, bulky grasping is employed in the aforementioned research work.

\begin{figure}[t!]
    \centering
    \includegraphics[scale = 1.17]{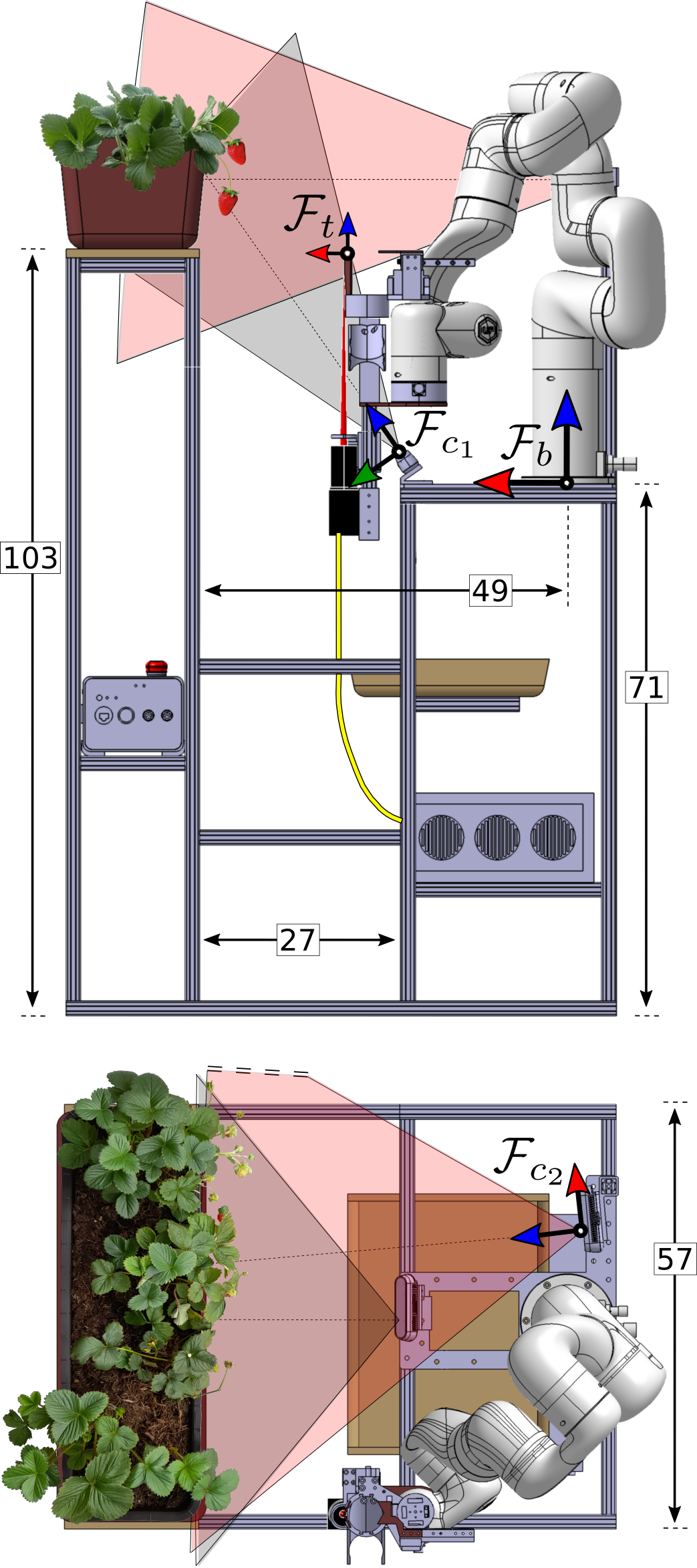}
    \captionsetup{aboveskip=10pt}
    \captionsetup{belowskip=-10pt}
    \caption{The harvesting system anatomy in side and top views. Relevant coordinate frames are shown with x, y, and z axes in red, green and blue arrows respectively. Vision cone of first and second RGB-D cameras in  grey and light red respectively. Selected dimensions of interest in centimeters.}
    \label{figure_setup_anatomy}
\end{figure}

\begin{figure}[b!]
    \centering
    \includegraphics[scale = 1.1]{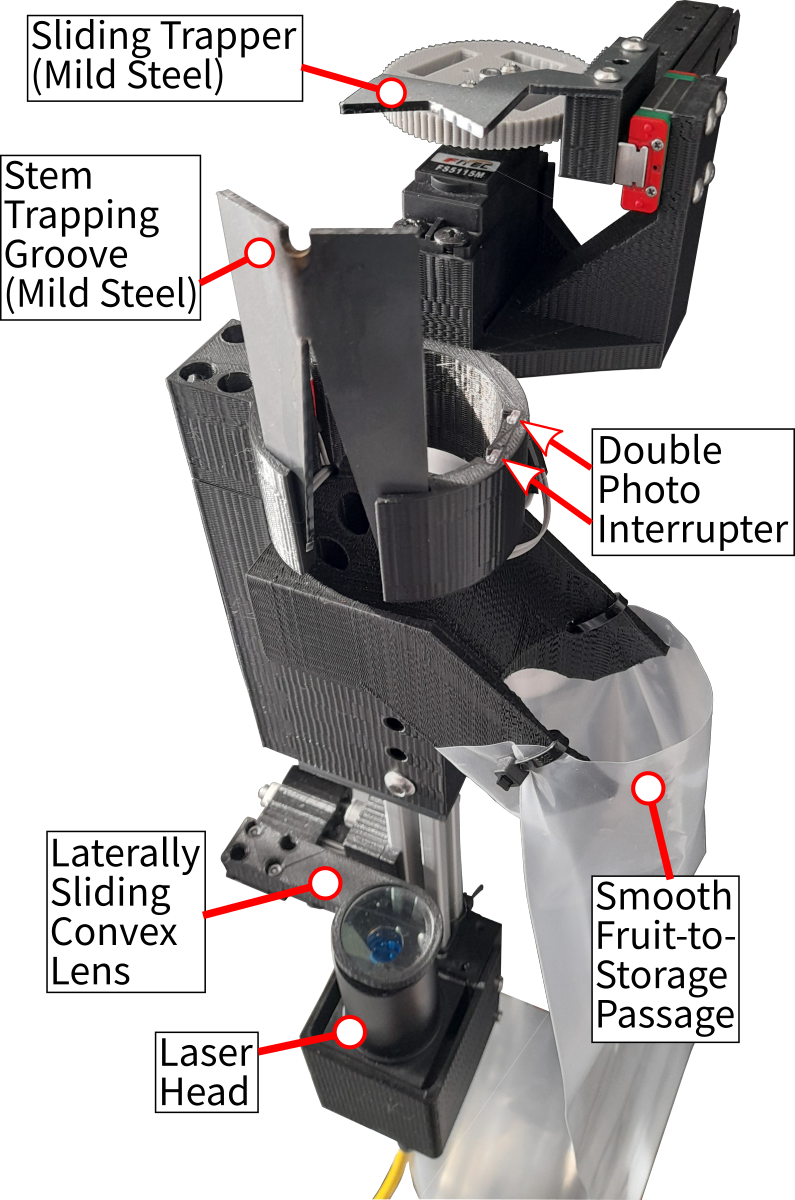}
    \captionsetup{aboveskip=10pt}
    \captionsetup{belowskip=-10pt}
    \caption{The harvesting tool anatomy with the process observing camera omitted for convenience.}
    \label{figure_harvesting_tool_anatomy}
\end{figure}

In this work, the authors present a novel harvesting tool that virtually captures the fruit by surrounding it, and detaches the stem by applying a highly focused laser beam from large distance, resulting in minimal interaction with the fruit and its local environment. In addition, the authors propose the harvesting system prototype shown in Fig. \ref{figure_setup_overview}, with relative distances provided in Fig. \ref{figure_setup_anatomy} replicating those normally found in strawberry growing poly-tunnel setups. The system as such, can be readily mounted to a mobile robot for field harvesting, with minimal modifications required. The contribution of our approach is threefold:
\begin{itemize}
    \item Productive: with average cycle time of $8$ seconds at $50$\% of maximum robot velocity, in addition to the maintenance-free laser cutting.
    \item Small footprint: interacting hardware width of $35mm$ greatly enhances fruit reachability.
    \item Robust: precise stem entrapment tolerating strawberry localization error of up to $\pm15mm$.
\end{itemize}
This paper is organised as follows, section II introduces the anatomy of the harvesting system and the operation logic. Algorithms for strawberry localization and harvesting as well as experiments are reported in section III. Conclusions are finally given in section IV.

\section{System Anatomy and Operation Logic}

The developed harvesting tool is shown in Fig. \ref{figure_harvesting_tool_anatomy}, two components interact with the strawberry namely; the stem trapping groove, and the stem trapper. Both are made of mild steel due to its low thermal conductivity (as compared to Aluminium for example), however, any ferrous metal will be as effective. When the laser beam is activated during the stem cutting process, this feature preserves the heat locally, helping the cutting process, killing plant viruses, and preventing the melt down of other plastic components of the tool. The width of both components is $35$ and $30mm$ respectively, matching the dimensions of a large strawberry, as such, minimum fruit dislocation can be achieved during harvesting, as well as better fruit reach-ability. The opening between both parts acts as the strawberry entry side in \cite{Sorour2022}, which will approach the fruit from below, completely surround it, before performing the stem cut. A convex lens with focal length of $25cm$ focuses the fiber laser beam generated at the laser head below \cite{fiber_laser}. It is servo controlled to perform a lateral, reciprocating, straight line motion to drive the focal point into a line with controllable length to effectively cut the trapped strawberry-stem. Two infrared photo interrupters are used to detect the free falling strawberry after being detached, that is then led to a storage unit by means of a smooth plastic passage, protecting the fruit surface from damage.

The harvesting tool is mounted to a cost-effective $6$ DOF collaborative robot arm system as shown in Fig. \ref{figure_setup_anatomy} in side and top views. A set of two RGB-D cameras mounted in different view points facing the strawberry growing trough, to the front and from below. The objective of such arrangement is to obtain a realistic point cloud of the scene with enough information to determine the location and the size of strawberries. The field of view of both cameras is depicted in the same figure, colored in light grey and red respectively, showing the common area of focus, at which strawberries can be mutually detected. This setup is based on an an average table-top height of $103cm$ readily available at the poly-tunnel of the host institution, and the average Thorvald mobile robot height \cite{GRIMSTAD_2017} widely used in field robotics. The robot arm base frame $\bm{\mathcal{F}}_{b}$ is related to the first $\bm{\mathcal{F}}_{c_1}$ and second $\bm{\mathcal{F}}_{c_2}$ camera frames by the constant transformation matrices ${^{b}\bm{\textbf{T}}_{c_1}}$, and ${^{b}\bm{\textbf{T}}_{c_2}}$ respectively, and rigidly connected via the arm base plate. As such, the setup (arm, tool, and cameras) can be fitted to a mobile robot for field testing with minimal  modification to the control algorithm. The tool tip frame $\bm{\mathcal{F}}_{t}$ is controlled to surround the fruit prior to harvesting, by manipulating the arm, it aligns with the tip of the stem trapping groove shown in Fig. \ref{figure_harvesting_tool_anatomy}.
Let the pose $^{b}\textbf{p}_{t} = \begin{bmatrix*} ^{b}\textbf{r}_{t}^{\top} & \!\!\! ^{b}\bm{\theta}_{t}^{\top} \end{bmatrix*}^{\top} \!\!\! \in \mathbb{R}^{6}$ of the manipulator's tool tip frame $\bm{\mathcal{F}}_{t}$ expressed in the base frame $\bm{\mathcal{F}}_b$ define the task space coordinates. With $^{b}\textbf{r}_{t} = \begin{bmatrix*} x & y & z \end{bmatrix*}^{\top}$ being the position vector, and $^{b}\bm{\theta}_{t} = \begin{bmatrix*} \gamma & \beta & \alpha \end{bmatrix*}^{\top}$ denoting a minimal representation of orientation (\textit{roll, pitch, yaw} RPY variation of Euler angles). In this work, the tool tip frame orientation is fixed and identical to that of the base frame, as such, the tool pose $^{b}\textbf{p}_{t}$ notation will be dropped, and only the $3D$ tool position $^{b}\textbf{r}_{t}$ will be used in the sequel. The arm configuration shown in Fig. \ref{figure_setup_anatomy} corresponds to the \texttt{HOME} tool position $^{b}\textbf{r}_{\texttt{HOME}}$, at which, the arm is not obstructing the field of vision of the cameras, and the strawberry localization algorithm (to be presented in the next section) can run.

\begin{figure}[t!]
    \centering
    \includegraphics[scale = 1.0]{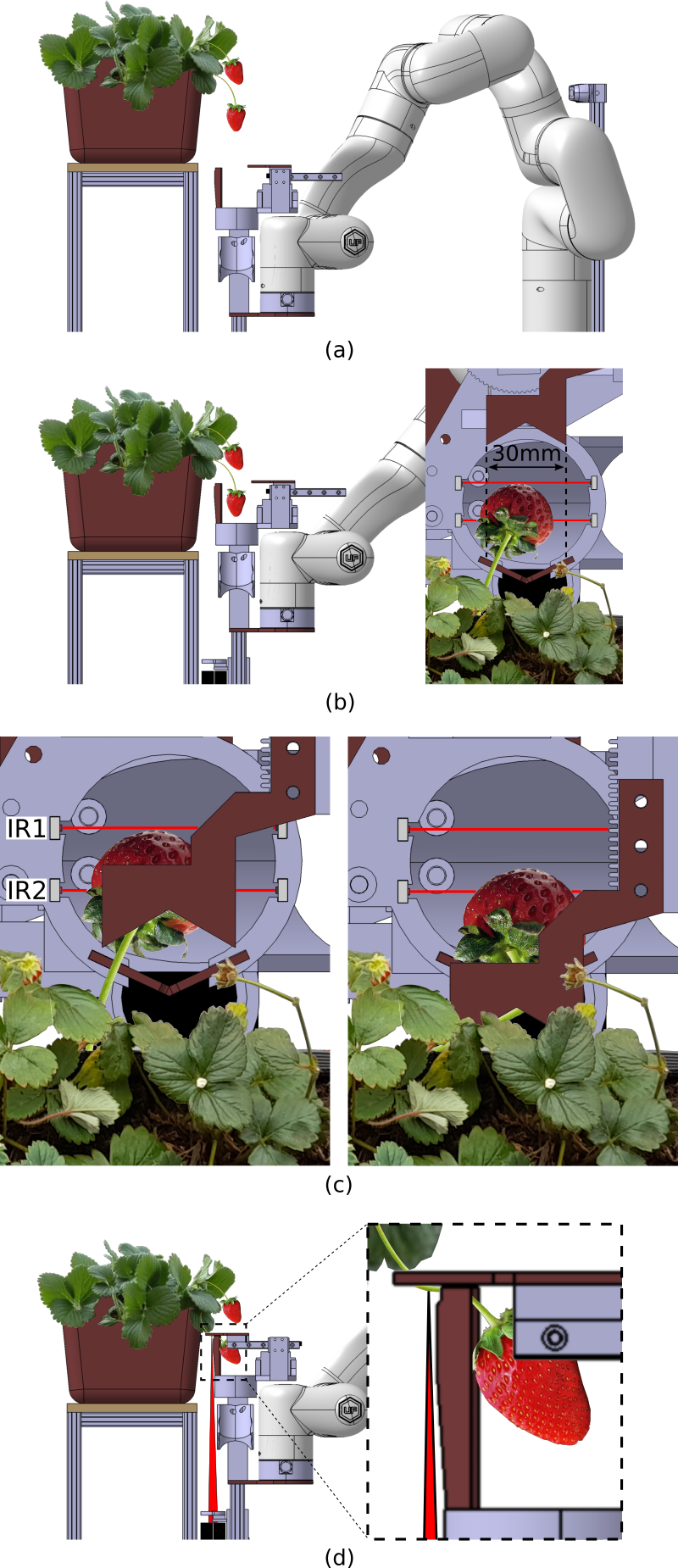}
    \captionsetup{aboveskip=13pt}
    \captionsetup{belowskip=-10pt}
    \caption{Operation logic, with the harvesting tool at the x-y coordinates of the strawberry in (a), eventually encapsulating it by moving upwards in z-axis (b). The stem is precisely entrapment into a groove in (c) while tolerating fruit localization error, followed by triggering the laser beam in (d) until fruit detachment.}
    \label{figure_operation_sequence}
\end{figure}

The operation logic of the harvesting tool is depicted in Fig. \ref{figure_operation_sequence}. Following strawberry localization, the v-shaped stem-trapping-groove (note the sliding trapper is also v-shaped, but inverted) - refer to Fig. \ref{figure_operation_sequence}(c)(left) - is positioned immediately behind the berry-to-harvest at a lower z-axis position in Fig\ref{figure_operation_sequence}(a), the tool tip then elevates to isolate the fruit from its surrounding to the rear as shown in Fig. \ref{figure_operation_sequence}(b). The trapper then slides to trap the stem, with Fig. \ref{figure_operation_sequence}(c)(left) showing the first contact between both, and the eventually fully trapped stem in \ref{figure_operation_sequence}(c)(right). The laser beam is then activated to full power as in Fig. \ref{figure_operation_sequence}(d), and the convex lens reciprocates laterally with a stroke of $6mm$ driving the focal point across the stem back and forth until the stem is cut. The mid-stroke is adjusted to the groove center point. Although the current tool design can force the stem into a precise location as compared to the author's previous work \cite{Sorour2022}, a laser beam with large focal distance is still needed to minimize the volume of the hardware interacting with the fruit during the harvesting process. The trapper end of stroke is adjusted to minimize the stem relocation, and as such the strawberry, while maintaining a metallic background shield as the laser beam performs the cut, this is shown in the enlarged square in \ref{figure_operation_sequence}(d).

The tool, aims to approach the berry from below while perfectly aligned with the trapper groove. However, due to inaccuracies resulting from the point cloud calculation, and augmenting the depth information from two different sensors, localization errors can add up, this in addition to the naturally occurring stem bending. The width of the trapper of $30mm$ ensures robustness against such errors within a tolerance of $\pm{15mm}$ from the actual strawberry location. This is shown in Fig. \ref{figure_operation_sequence}(b) and Fig. \ref{figure_operation_sequence}(c)(left), with imperfect strawberry localization. Fruit detachment is detected using two photo interrupters, the infrared beam of which is virtually shown in red color in Fig. \ref{figure_operation_sequence}(b) and \ref{figure_operation_sequence}(c) for clarification. Once any of the two beams is interrupted, the laser source is disengaged, terminating the cutting cycle, and the arm either moves to the next fruit to cut, or to the home position.

\begin{figure}[t!]
    \vspace{-7pt}
    \centering
    \includegraphics[scale = 0.25]{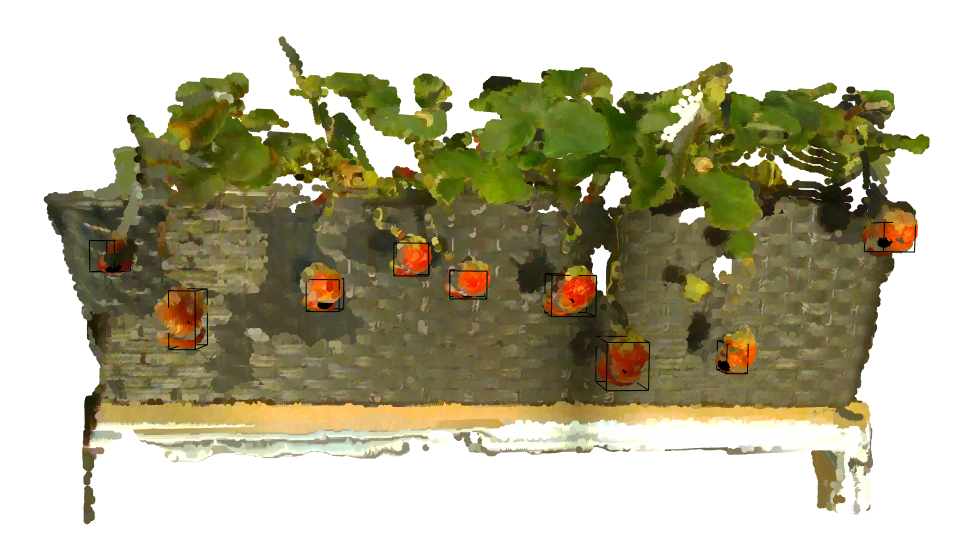}
    \captionsetup{aboveskip=10pt}
    \captionsetup{belowskip=-10pt}
    \caption{Reduced scene point cloud focusing on the area of interest with black bounding box marking the located strawberries.}
    \label{figure_point_cloud_scene_front}
\end{figure}

\begin{algorithm}[b!]
 \caption{Strawberry localization algorithm}
 \begin{algorithmic}[1]
 \renewcommand{\algorithmicrequire}{\textbf{Input:}}
 \renewcommand{\algorithmicensure}{\textbf{Output:}}
 \REQUIRE colored scene point cloud $^{c_1}\mathcal{C}_{s_1}$ from RGB-D CAM1, \\
          \hspace{4.5mm} colored scene point cloud $^{c_2}\mathcal{C}_{s_2}$ from RGB-D CAM2. \\
 \ENSURE  Vectors defining the bounding box for localized strawberries in arm base frame.
  \STATE $^{b}\mathcal{C}_{s} = {^{b}\bm{\textbf{T}}_{c_1}} ^{c_1}\mathcal{C}_{s_1} + {^{b}\bm{\textbf{T}}_{c_2}} ^{c_2}\mathcal{C}_{s_2} $
  \STATE $^{b}\mathcal{C}_{rs} = \emptyset$
  \FORALL{point $^{b}{c}^{i}_{s}$ in $^{b}\mathcal{C}_{s}$}
   \IF {($^{b}{c}^{i}_{s}.x < x^{+}_{l}$ \textbf{and} $^{b}{c}^{i}_{s}.x > x^{-}_{l}$ \textbf{and} $^{b}{c}^{i}_{s}.y < y^{+}_{l}$ \textbf{and} $^{b}{c}^{i}_{s}.y > y^{-}_{l}$ \textbf{and} $^{b}{c}^{i}_{s}.z < z^{+}_{l}$ \textbf{and} $^{b}{c}^{i}_{s}.z > z^{-}_{l}$ )}
   \STATE $^{b}\mathcal{C}_{rs} \gets {^{b}{c}^{i}}_{s}$
   \ENDIF
  \ENDFOR
  \STATE $^{b}\mathcal{C}_{red} = \emptyset$
  \FORALL{point $^{b}{c}^{i}_{rs}$ in $^{b}\mathcal{C}_{rs}$}
   \IF {($^{b}{c}^{i}_{rs}.r > r_{th}$ \textbf{and} $^{b}{c}^{i}_{rs}.g < g_{th}$ \textbf{and} $^{b}{c}^{i}_{rs}.b < b_{th}$ )}
    \STATE $^{b}\mathcal{C}_{red} \gets {^{b}{c}^{i}_{rs}}$
   \ENDIF
  \ENDFOR
  \STATE $^{b}\mathcal{C}_{straw} = \texttt{EuCS(}t, s_{min}, s_{max}, ^{b}\mathcal{C}_{red} \texttt{)}$
  \FORALL{cluster $^{b}\textbf{c}^{i}_{straw}$ in $^{b}\mathcal{C}_{straw}$}
    \STATE $^{b}\textbf{x}_{straw}^{min} \gets \texttt{MIN(} ^{b}\textbf{c}^{i}_{straw}, \texttt{0)}$
    \STATE $^{b}\textbf{x}_{straw}^{max} \gets \texttt{MAX(} ^{b}\textbf{c}^{i}_{straw}, \texttt{0)}$
    \STATE $^{b}\textbf{y}_{straw}^{min} \gets \texttt{MIN(} ^{b}\textbf{c}^{i}_{straw}, \texttt{1)}$
    \STATE $^{b}\textbf{y}_{straw}^{max} \gets \texttt{MAX(} ^{b}\textbf{c}^{i}_{straw}, \texttt{1)}$
    \STATE $^{b}\textbf{z}_{straw}^{min} \gets \texttt{MIN(} ^{b}\textbf{c}^{i}_{straw}, \texttt{2)}$
    \STATE $^{b}\textbf{z}_{straw}^{max} \gets \texttt{MAX(} ^{b}\textbf{c}^{i}_{straw}, \texttt{2)}$
  \ENDFOR
 \RETURN $^{b}\textbf{x}_{straw}^{min}, {^{b}\textbf{x}_{straw}^{max}}, {^{b}\textbf{y}_{straw}^{min}}, {^{b}\textbf{y}_{straw}^{max}},$\\ \hspace{10.8mm} $^{b}\textbf{z}_{straw}^{min}, {^{b}\textbf{z}_{straw}^{max}}$
 \end{algorithmic}
\label{strawberry_localization_algorithm}
\end{algorithm}

\begin{figure*}[t!]
    \centering
    \includegraphics[width=1.0\textwidth]{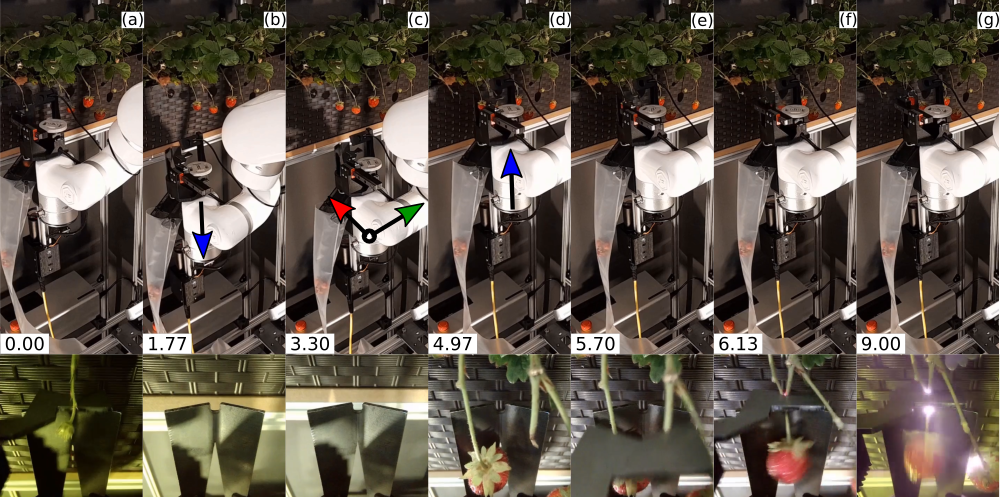}
    \captionsetup{aboveskip=10pt}
    \captionsetup{belowskip=-10pt}
    \caption{Snapshots of a complete strawberry harvesting cycle, starting and ending with the detachment of a single strawberry, indicating the time stamp (in seconds) of each step in-between. Lower row showing the close-up images captured by the tool-observing camera.}
    \label{figure_harvesting_demo_snapshots}
\end{figure*}

\section{Control Algorithm and Experiments}

In this section we present the algorithms for strawberry localization and motion control that are used in the harvesting experiment. In which, the system described in the previous section is used to harvest a collection of $9$ strawberries in full autonomy, the point cloud of these depicted in Fig. \ref{figure_point_cloud_scene_front}. The hardware used feature the $6$ DOF xARM6 collaborative robot \cite{xarm}, two realsense D435 depth cameras, and a $50$ Watts Raycus RFL-P50QB module \cite{fiber_laser} as the fiber laser source. The harvesting tool is 3D printed except for the stem-trapper and the trapping-groove that are subjected to extensive laser heat. Microcontroller is used to control the convex lens and trapper movements, the laser activation, as well as monitoring the strawberry detachment photo interrupters. On the software side, we use the realsense SDK library for interfacing with the cameras, and the Point Cloud Library (PCL) \cite{Rusu_ICRA2011_PCL}. A video of the harvesting experiment is submitted with this paper, also available on YouTube\footnote{\url{https://youtu.be/W3UyDt_7erA}.}, snapshots of which are shown in Fig. \ref{figure_harvesting_demo_snapshots}.


\subsection{Strawberry Localization}

The pseudo-code for strawberry localization is provided in Algorithm \ref{strawberry_localization_algorithm}, it takes as input the raw scene point cloud acquired by both RGB-D cameras $^{c_1}\mathcal{C}_{s_1}$, and $^{c_2}\mathcal{C}_{s_2}$ expressed in the corresponding camera frame. These are then transformed to the arm base frame and augmented (code line 1) to form the scene point cloud $^{b}\mathcal{C}_{s}$. From which, a \textit{reduced scene} cloud set $^{b}\mathcal{C}_{rs} \subset {^{b}\mathcal{C}}_{s}$ is then constructed, where a point ${^{b}{c}^{i}}_{s}$ in the scene cloud $^{b}\mathcal{C}_{s}$ is added to the reduced cloud if it resides in a spatial window characterized by the maximum and minimum limits $x^{+}_{l}, x^{-}_{l}, y^{+}_{l}, y^{-}_{l}, z^{+}_{l}, z^{-}_{l}$ in $x, y,$ and $z$ coordinates respectively of the arm base frame. This reduces the forthcoming computation to the area of interest that is dexterously accessible by the robot arm, such reduced point cloud is shown in Fig. \ref{figure_point_cloud_scene_front}. Ripe strawberries are simply extracted by thresholding the red color using the RGB thresholds $r_{th}, g_{th}, b_{th}$ to form the \textit{red} point cloud $^{b}\mathcal{C}_{red} \subset {^{b}\mathcal{C}}_{rs}$. We then use the PCL implemented Euclidean cluster segmentation algorithm to segment each individual strawberry. The output of the function $\texttt{EuCS(}t, s_{min}, s_{max}, ^{b}\mathcal{C}_{red} \texttt{)}$ is a set of strawberry clusters (set of point clouds each representing a single strawberry) $^{b}\mathcal{C}_{straw}$ arranged in ascending order of the y-axis coordinate values. With $t, s_{min}, s_{max}$ denoting the segmentation tolerance, minimum and maximum cluster sizes respectively.  The $\texttt{MIN(} ^{b}\textbf{c}^{i}_{straw}, \texttt{0)}$ function, supplied with a point cloud $^{b}\textbf{c}^{i}_{straw}$ and an index, will return the minimum value available at such index in all point cloud points, whereas, supplied with a vector, will return the smallest value irrespective of the index. The lengthy pseudo-code for the aforementioned two functions is omitted for convenience. Values of the parameters used in Algorithm \ref{strawberry_localization_algorithm} is provided in Table \ref{table_localization_algorithm_parameters}. The output is a set of vectors defining the bounding boxes of the localized strawberries, these boxes are shown in Fig. \ref{figure_point_cloud_scene_front}.

\begin{table}[b]
\captionsetup{aboveskip=-1pt,belowskip=-10pt}
\caption{Localization algorithm parameters}
\label{table_localization_algorithm_parameters}
\begin{center}
\begin{tabular}{|l|c||l|c|}
\hline
Parameter & Value(unit) & Parameter & Value(unit) \\
\hline
\hline

$x^{+}_{l}$  & $55(cm)$  & $r_{th}$     & $100$ \\
$x^{-}_{l}$  & $25(cm)$  & $g_{th}$     & $70$ \\
$y^{+}_{l}$  & $30(cm)$  & $b_{th}$     & $70$ \\
$y^{-}_{l}$  & $-30(cm)$ & $t$          & $0.02(cm)$ \\
$z^{+}_{l}$  & $50(cm)$  & $s_{min}$    & $20$ \\
$z^{-}_{l}$  & $30(cm)$  & $s_{max}$    & $1000$ \\

\hline
\end{tabular}
\end{center}
\end{table}

\subsection{Harvesting Motion}

\begin{algorithm}[b!]
 \caption{Harvesting algorithm}
 \begin{algorithmic}[1]
 \renewcommand{\algorithmicrequire}{\textbf{Input:}}
 \renewcommand{\algorithmicensure}{\textbf{Output:}}
 \REQUIRE $^{b}\textbf{x}_{straw}^{max}, {^{b}\textbf{y}_{straw}^{min}}, {^{b}\textbf{y}_{straw}^{max}}, {^{b}\textbf{z}_{straw}^{min}}, {^{b}\textbf{z}_{straw}^{max}} $.  \\
 \ENSURE  desired tool tip position in arm base frame $^{b}\textbf{r}_{t}^{d}$.
  \STATE $\texttt{ROBOT\_MOVE(} ^{b}\textbf{r}_{\texttt{HOME}} \texttt{)} $
  \STATE $^{b}{z}_{min} = \texttt{MIN(} ^{b}\textbf{z}_{straw}^{min} \texttt{)} - \texttt{10mm}$
  \FORALL{strawberry unit \texttt{i} in $n_{straw}$ }
   \STATE $^{b}\textbf{r}_{t}^{d}\texttt{(2)} = {^{b}{z}_{min}}$
   \STATE $\texttt{ROBOT\_MOVE(} ^{b}\textbf{r}_{t}^{d} \texttt{)} $
   \STATE $^{b}\textbf{r}_{t}^{d}\texttt{(0)} = {^{b}\textbf{x}_{straw}^{max}\texttt{(i)}} + \texttt{10mm}$
   \STATE $^{b}\textbf{r}_{t}^{d}\texttt{(1)} = ({^{b}\textbf{y}_{straw}^{min}\texttt{(i)}} + {^{b}\textbf{y}_{straw}^{max}\texttt{(i)}})/2$
   \STATE $\texttt{ROBOT\_MOVE(} ^{b}\textbf{r}_{t}^{d} \texttt{)} $
   \STATE $^{b}\textbf{r}_{t}^{d}\texttt{(2)} = {^{b}\textbf{z}_{straw}^{max}\texttt{(i)}} + \texttt{15mm}$
   \STATE $\texttt{ROBOT\_MOVE(} ^{b}\textbf{r}_{t}^{d} \texttt{)} $
   \STATE \texttt{TRAP\_STEM()}
   \STATE \textbf{while} \texttt{IR1} \textbf{and} \texttt{IR2} \textbf{do}
   \STATE \hspace{2.7mm} \texttt{LASER(ON)}
   \STATE \textbf{end while}
   \STATE \texttt{LASER(OFF)}
   \STATE \texttt{RELEASE\_STEM()}
  \ENDFOR
  \STATE $\texttt{ROBOT\_MOVE(} ^{b}\textbf{r}_{\texttt{HOME}} \texttt{)} $
 \end{algorithmic}
\label{harvesting_algorithm}
\end{algorithm}

Pseudo-code of the harvesting algorithm is provided in Algorithm \ref{harvesting_algorithm}, it is also depicted in Fig. \ref{figure_harvesting_demo_snapshots}. It takes as input, the vectors defining the  bounding box of the identified/localized strawberries, that is the output of the localization algorithm in Algorithm \ref{strawberry_localization_algorithm}. The output is a series of commands to control the robot arm movement, as well as controlling the stem-trapper and the laser module. In this work, a strawberry cutting cycle, and as such the cycle time, starts and ends with a single strawberry unit being detached. Initially, the algorithm moves the robot arm to the home position $\textbf{r}_{\texttt{HOME}}$ (configuration depicted in Fig. \ref{figure_setup_anatomy}), the $\texttt{ROBOT\_MOVE()} $ function in Algorithm \ref{harvesting_algorithm} is blocking (has to be finished before executing the next line of code). It then computes a minimal z-axis point $^{b}{z}_{min}$, corresponding to the lowest hanging strawberry observed. In Fig. \ref{figure_harvesting_demo_snapshots}(a), a strawberry unit has just been cut, signaling the start of a new cycle, at which the robot arm moves to $^{b}{z}_{min}$, as shown in Fig. \ref{figure_harvesting_demo_snapshots}(b) consuming $1.77$ seconds. By setting $^{b}\textbf{r}_{t}^{d}\texttt{(2)} = {^{b}{z}_{min}}$, only the z-axis component of the desired tool position vector $^{b}\textbf{r}_{t}^{d}$ is changed, while retaining the old values for the other two components (x and y axes). The arm then moves to the x-y coordinates of the next detected strawberry using the information of the corresponding bounding box, before moving upwards in z-axis as shown in Fig. \ref{figure_harvesting_demo_snapshots}(c) and Fig. \ref{figure_harvesting_demo_snapshots}(d) respectively. In order to compensate for the probable inaccuracy in measuring the actual depth of strawberry, an empirically determined safety factor is added to the maximum boundary computed in x and z axes (code lines 6 and 9 in Algorithm \ref{harvesting_algorithm}). To this end, the robot arm motion for this cycle is finished, what remains is the cutting procedure, that is microcontroller implemented. When given the signal to perform the cut, the microcontroller actuates the trapper forward using \texttt{TRAP\_STEM()}, then activates the laser module as well as the convex lens lateral movement using \texttt{LASER(ON)} until any of the photo interrupters \texttt{IR1} or \texttt{IR2} (refer to Fig. \ref{figure_operation_sequence}) detects the detached fruit. At which instance, the laser beam is deactivated and the trapper moves backwards.

The average cycle and stem-cut time is $8.02$, and $2.3$ seconds respectively, with the robot arm operating at $50$\% of maximum velocity for hardware safety reasons. The localization algorithm consumes $100ms$ in worst case scenario on a standard laptop with intel core-i$7$ processor. The maintenance-free laser cutting has a long term positive impact on productivity, as compared to literature-dominating mechanical cutting tools that require frequent replacement. To this end, the authors believe that a $3D$ linear system replacing the robot arm, in addition to a $100$ Watt laser module (following the conclusions in \cite{Sorour2022_IROS}) can reduce the average cycle time below the $4$ seconds mark, cutting the gap with the average manual picking cycle standing at $1.5$s \cite{Woo2020}. That, in addition to further enhancing the fruit identification and localization algorithm, would greatly update the current development status \cite{Zhou2022} thanks to the minimal footprint of the developed tool.

\section{Conclusion}
In this work, a novel harvesting prototype has been presented, customised for table-top grown strawberries. It employs a trapping mechanism to surround the fruit and force the stem into a pre-defined position where a focused laser beam performs the cut. The proposed harvesting tool is robust against localization errors while maintaining small footprint for enhancing fruit reachability. Successful harvesting demonstration confirm the effectiveness of the novel methodology.


\bibliographystyle{IEEEtran}  
\bibliography{Bib}

\end{document}